\documentclass[conference]{IEEEtran}
\IEEEoverridecommandlockouts
\usepackage{cite}
\usepackage{amsmath,amssymb,amsfonts}
\usepackage{algorithmic}
\usepackage{graphicx}
\usepackage{textcomp}
\usepackage{xcolor}
\def\BibTeX{{\rm B\kern-.05em{\sc i\kern-.025em b}\kern-.08em
    T\kern-.1667em\lower.7ex\hbox{E}\kern-.125emX}}

\usepackage{amsmath}

\usepackage{multirow}

\usepackage{amsmath}
\usepackage{enumitem}
\setlist{leftmargin=3mm}
\usepackage{amssymb}

\usepackage{algorithm}
\usepackage{algorithmic}
\usepackage{url}
\usepackage{threeparttable}

\usepackage{hyperref}
\hypersetup{breaklinks=true}
\usepackage[hyphenbreaks]{breakurl}

\usepackage{graphicx}
\usepackage{color} 

\begin{document}

\title{FedKNOW: Federated Continual Learning with Signature Task Knowledge Integration at Edge\\
%
\thanks{This work is supported by the National Natural Science Foundation of China (Grant No. 61872337, 62132019, 62272046) and Shandong Provincial Natural Science Foundation (Grant No. ZR2020MF034). Corresponding author: Rui Han.}
}

\author{\IEEEauthorblockN{Yaxin Luopan, Rui Han, Qinglong Zhang, Chi Harold Liu, Guoren Wang}
\IEEEauthorblockA{\textit{Beijing Institute of Technology, Beijing, China} \\
Beijing, China \\
\{1120181200, hanrui, 3120211050, chiliu, wanggr\}@bit.edu.cn}
\and
\IEEEauthorblockN{Lydia Y. Chen}
\IEEEauthorblockA{\textit{TU Delft} \\
Delft, Netherlands \\
\{lydiaychen@ieee.org}
}

\maketitle

\begin{abstract}

Deep Neural Networks (DNNs) have been ubiquitously adopted in internet of things and are becoming an integral of our daily life. When tackling the evolving learning tasks in real world, such as classifying different types of objects, DNNs face the challenge to continually retrain themselves according to the tasks on different edge devices. Federated continual learning is a promising technique that offers partial solutions but yet to overcome the following difficulties: the significant accuracy loss due to the limited on-device processing, the negative knowledge transfer caused by the limited communication of non-IID data, and the limited scalability on the tasks and edge devices. In this paper, we propose FedKNOW, an accurate and scalable  federated continual learning framework, via a novel concept of signature task knowledge.
\textcolor{black}{FedKNOW is a client side solution that continuously extracts and integrates the knowledge of signature tasks which are highly influenced by the current task. Each client of FedKNOW is composed of a knowledge extractor, a gradient restorer and, most importantly, a gradient integrator. Upon training for a new task, the gradient integrator ensures the prevention of \emph{catastrophic forgetting} and mitigation of \emph{negative knowledge transfer} by effectively combining signature tasks identified from the past local tasks and other clients' current tasks through the global model.}
We implement FedKNOW in PyTorch and extensively evaluate it against state-of-the-art techniques using popular federated continual learning benchmarks. Extensive evaluation results on heterogeneous edge devices show that FedKNOW improves model accuracy by 63.24\% without increasing model training time, reduces communication cost by 34.28\%, and achieves more improvements under difficult scenarios such as large numbers of tasks or clients, and training different complex networks.

\end{abstract}




\section{Introduction} \label{Section: Introduction}

\begin{figure} [htp]
\centering
  \includegraphics[scale=0.48]{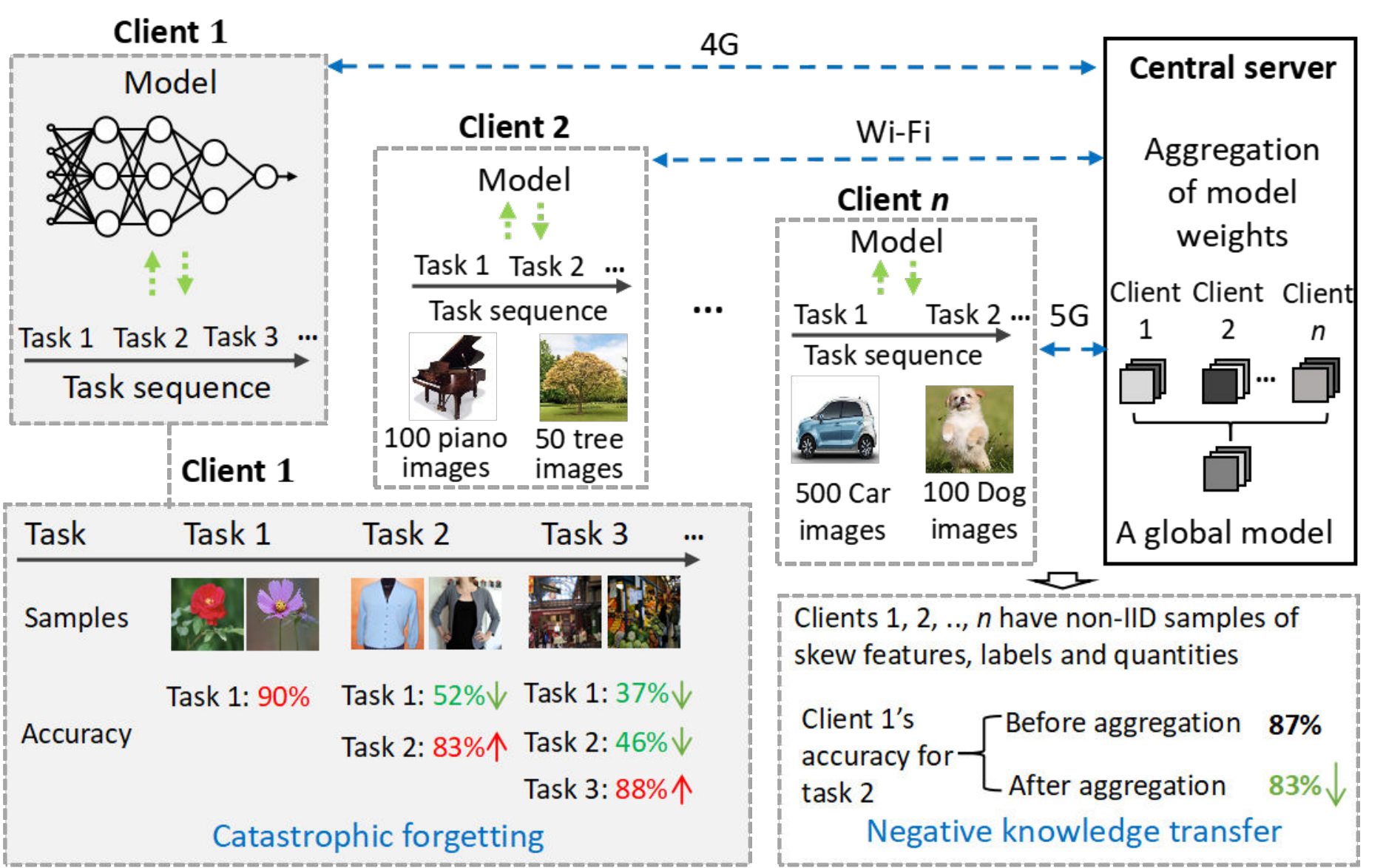}\\
  \caption{\textcolor{black}{An example scenario of federated continual learning with $n$ clients}}
  \label{Fig: Example}
\end{figure}

Today, billions of mobile and Internet of Things (IoT) devices generate zillions bytes of data at the network edge, offering opportunities to deploy artificial intelligence (AI) locally on edge devices\footnote{We also refer to such edge devices as clients.}. Such on-device AI applications, e.g. deep neural networks (DNNs), have the advantage of avoiding transmitting raw data and hence preserving data privacy~\cite{zhou2019edge}.
At the same time, the arising new challenge is that the environment continuously evolves, requiring the DNN models to retrain and adapt to those changes~\cite{chen2019deep}.
For example, Figure \ref{Fig: Example} illustrates the DNN model in client 1 needs to \textcolor{black}{handle a sequence of tasks (e.g. image classification or object detection) over time. Typically, a \textbf{task} is composed of multiple classes/objects (e.g. different animals or vehicles) and different features for each class~\cite{parisi2019continual}.}



\textbf{Federated continual learning (FCL)}. Continual learning is a prevalent technique that incrementally learns deep models from such a non-stationary data consisting of different tasks. Traditional continual learning only learns its models from the training samples on their hosted devices. In contrast, humans can learn from their own and others' past experiences through conversations, lectures, books and other means.
Motivated by the intuition of learning from other clients' (indirect) experience, FCL combines continual learning in the federated learning framework such that the model in a client can continuously learn from its local data and the knowledge of the tasks in other clients~\cite{yang2019federated,yoon2021federated}.
As shown in the exemplary scenario of Figure \ref{Fig: Example}, a central server obtains the model weights/parameters locally trained in $n$ clients, aggregates them into a global model, and sends it back to all clients. This allows each client to perform continual learning of its task sequence based on its local data, while learning from other clients by communicating their task-specific weights via the server.
\textcolor{black}{One major problem of performing continual learning in a client is \textbf{catastrophic forgetting}: when its model learns new tasks over time, it may forget previously learned task information and the model accuracy in these tasks degrades~\cite{robins1995catastrophic,grossberg2012studies,de2021continual}.
The challenge of coping with evolving task is further exacerbated when training DNN models according to evolving tasks on a large number of clients.
In FCL, each client has its private sequence of tasks. Even for the same task, different clients host \emph{non-IID datasets} whose distributions of classes, input data features, and numbers of samples vary~\cite{zhu2021federated}. This means although starting from the same global model, these clients have diverse models after local training. When aggregating a client¡¯s model with those of other clients, the model divergence may decrease the accuracy in local tasks, know as \textbf{negative knowledge transfer}~\cite{karimireddy2019scaffold,deng2020adaptive,zhao2018federated,li2019convergence,zhu2021federated}.}







\textbf{Challenges of FCL at edge}. \textcolor{black}{Edge computing is developed to reduce communication costs to cloud servers and enhance data privacy via on-device data processing~\cite{satyanarayanan2017emergence,shi2016edge}.
Performing FCL at edge brings new problems such that the computation and communication costs in model training increase with the number of tasks and client, and conducting such expensive training on resource-constrained edge devices give rise to two technical challenges.
}

\emph{Limited computational resources lead to significant accuracy losses}. Existing continual learning and FCL techniques are server-side solutions, which designed for powerful cloud servers and retain training samples or model weights of previous tasks from all clients to avoid catastrophic forgetting ~\cite{lopez2017gradient,chaudhry2018efficient,raghavan2021formalizing,cha2021co2l,kirkpatrick2017overcoming,aljundi2018memory,jung2020continual,yao2020continual,yoon2021federated}. This means the learning process becomes longer when the number of tasks increases. For example, the training time of a ResNet-18~\cite{he2016deep} increases by 50 times when the number of tasks increases from 1 to 80.
When encountering resource constraints, these techniques can only retain a portion of samples and may incur large accuracy losses (20\% to 50\% losses) because most of important information in previous tasks is dropped.
The \textbf{first challenge}, therefore, is to design a lightweight learning method that can keep extensive historical knowledge and take short model training time directly on resource-constrained edge devices.



\emph{Preventing negative knowledge transfer causes high communication costs and privacy leakage}.
Existing techniques rely on the central server which collects and keeps all clients' task models to prevent negative knowledge transfer~\cite{yao2020continual,yoon2021federated}.
This mechanism causes high communication costs because: (i) the knowledge's size increases linearly with the number of clients; and (ii) all clients need to synchronize other clients' latest knowledge once any new task arrives.
For instance, the communication traffic of FedWEIT~\cite{yoon2021federated} is eight times larger than that of the basic federate learning method when the client number is just 20.
\textcolor{black}{Maintaining the global knowledge among multiple clients also violates scalability and privacy enforcement of edge computing.
The \textbf{second challenge} is how to develop a distributed method that can prevent negative knowledge transfer without increasing extra communications among clients.}



\textcolor{black}{In this paper, we depart from computationally and communication intensive FCL server-side approaches and propose FedKNOW, a lightweight client-side solution that integrates knowledge of signature tasks which encompass the relevant past and peer tasks.
FedKNOW acts in each client and extracts compact and transferable knowledge (instead of data) -- the critical subset of model weights.
When learning a new task, FedKNOW integrates it with the knowledge of its \textbf{signature tasks}, which are the new task's most dissimilar tasks identified from local past tasks to prevent \emph{catastrophic forgetting}, and the updated global model representing other clients' current tasks in preventing \emph{negative knowledge transfer}.
By completing knowledge integration with polynomial time complexity, FedKNOW addresses the limitations of existing techniques by providing both high model accuracy and low communication overhead at edge.} In particular, the contributions of this paper are as follows:






\textbf{Scalable client-side solution through the knowledge of signature tasks}. \textcolor{black}{In contrast the the prior art, FedKNOW is a client side method that acts on the knowledge of signature tasks, resulting into light-weight computation and communication for clients. FedKNOW extracts and retains each task's \emph{knowledge} as a small proportion (e.g. 10\%) of model weights with the highest degree of activation to the task.}

\textbf{High-accuracy model training via gradient integration}. \textcolor{black}{When learning a new task in a client, FedKNOW designs an optimization approach that integrates its gradient with gradients of previously experienced tasks, and integrates its gradients before and after global aggregation. Both integrations guarantee the acute angle between the integrated gradient and all other gradients. This gradient is then used in model updating to prevent catastrophic forgetting and negative knowledge transfer.}




\textbf{Convergence proof and evaluation}. We prove the convergence of FedKNOW under the constraints of learning rates. We also fully implement FedKNOW on top of PyTorch to support deep learning applications on edge devices, and conduct extensive evaluation against the state-of-the-art techniques, i.e. continual learning, federated learning and federated continual learning, using popular continual learning benchmarks.

\emph{Summary of experimental results.}
\emph{(i) Extensible in terms of architecture}. We test FedKNOW on five types of heterogeneous edge devices (Jetson TX2, Nano, Xavier NX, AGX, and Raspberry Pi).
\emph{(ii) High-accuracy continual learning}. We extensively compare FedKNOW to the 11 state-of-the-art techniques, and find that it increases model accuracy by 63.24\% using similar or less model training time, for extremely challenging scenarios of 80 tasks or 100 clients. Compared to standard continual learning that uses all the data sample, FedKNOW achieves higher accuracy because of effective avoidance of negative knowledge transfer.
\emph{(iii) Low-overhead federated model training}. To complete the same model training jobs, FedKNOW reduces communication time by 34.28\% compared to the latest FCL technique, especially FedWEIT.
\textcolor{black}{\emph{(iv) Applicability to different scenarios and DNNs.} We ensure that our approach works well when the numbers of tasks and clients increase, and the condition of communication changes. We also demonstrate the applicability of FedKNOW on 8 prevalent DNNs~\cite{khan2020survey}.
}



\section{Related Work} \label{Section: Relatedwork}

The major problem faced by federated continual learning is the catastrophic forgetting in neural networks when learning new tasks~\cite{robins1995catastrophic}. This problem is further complicated by the negative transfer in federated learning due to the Non-IID data sets in different clients~\cite{li2019convergence}.
Here are existing techniques designed to address these problems.

\textbf{Continual learning}. Mainstream techniques designed to address catastrophic forgetting can be divided into three categories:
(1) \emph{memory rehearsal} uses a memory cache to store the samples of previous tasks and use them in learning the new task to avoid forgetting. Hence their computational costs increase with the number of tasks~\cite{lopez2017gradient,chaudhry2018efficient,raghavan2021formalizing}.
(2) \emph{Regularization-based techniques} estimate different parameters/weights' contributions to a model and maintain part of information in important weights when learning new tasks~\cite{kirkpatrick2017overcoming,cha2021co2l,aljundi2018memory,jung2020continual}.
(3) \emph{Dynamic architectural techniques} design different models for different tasks and solve catastrophic forgetting by isolating part of the model parameters~\cite{rusu2016progressive,yoon2017lifelong,li2017learning}. However, it is difficult to apply these techniques in federated learning, because they require massive retained samples or weights to increase model generalization/accuracy, while restricting to learn tasks from the local data without benefiting from other clients' knowledge.

\textbf{Federated learning} trains a global model using private data from multiple clients~\cite{li2020federated,li2021efficient}.
In a client's model training, negative transfer from other clients (due to their Non-IID datasets) is a crucial problem that delays training convergency and degrades model accuracy~\cite{li2019convergence}.
\emph{Personalized federated learning} is a major technique used to mitigate such negative transfer and it can be divided into four types: (1) Mixture model techniques such as adaptive personalized federated learning (APFL)~\cite{deng2020adaptive} dynamically change the ratio of global and local models in training.
(2) In local fine-tuning techniques, each client first accepts a global model, and then updates it using local data. Meta-learning~\cite{finn2017model,chen2018federated, khodak2019adaptive} is increasingly employed to complete the update within a few iterations. 
(3) Contextualization aims to provide a different model for each context, e.g., character's context ~\cite{wang2019federated}. 
(4) Multi-task learning lets each client train a separate task~\cite{smith2017federated} and further classifies clients into different groups according to their tasks~\cite{kairouz2021advances}.
Note that the last two types of techniques cannot be directly applied in continual learning, because they compound learning new tasks with contexts and multi-task learning.


\textbf{FCL}. Some initial technique proposes server-side solution that maintains some training samples in the server and uses them in global model updating to avoid catastrophic forgetting~\cite{yao2020continual}. The effectiveness of this work depends on the amounts of maintained samples and it is impractical to share clients' local data in the server due to data privacy~\cite{mcmahan2017communicationefficient}.
The latest FCL technique, FedWEIT~\cite{yoon2021federated} uses adaptive model weights to maintain previous tasks' knowledge in a client and retains all clients' adaptive weights in the central server. Whenever a client needs to learn a new task, it first obtains the server's adaptive weights and then uses them in model training to prevent both catastrophic forgetting and negative transfer. The main limitation of FedWEIT is the scalability with respect to the number of clients and tasks due to their communication overhead. 


\section{Design of FedKNOW} \label{Section: FedKNOW}


\subsection{Overview} \label{Section: Overview}


\begin{figure*}[t!]
\centering
  \includegraphics[scale=0.51]{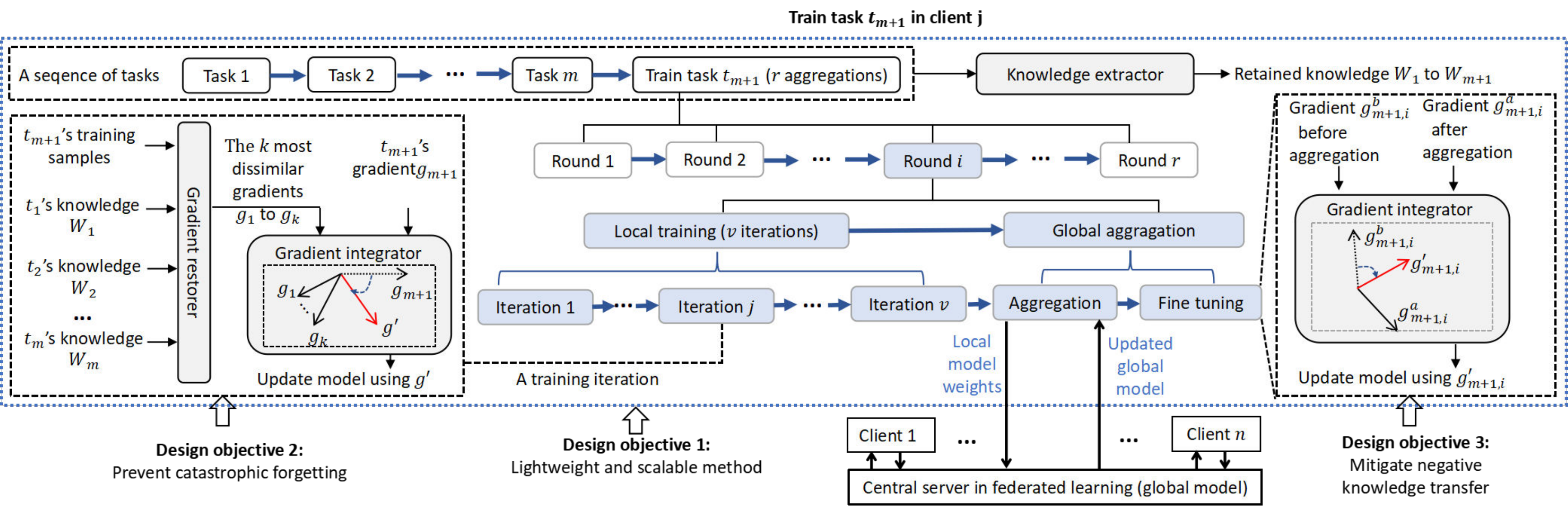}\\
  \caption{\textcolor{black}{FedKNOW process and its three design objectives}}
  \label{Fig: Overview}
\end{figure*}

\textcolor{black}{We design FedKNOW to continually train sequences of different learning tasks on federated clients. FedKNOW features on a novel concept of signature task knowledge which further enables lightweight computation and communication on resource-constrained edge devices. As shown in Figure \ref{Fig: Overview}, FedKNOW acts in each federated client and is composed of three components: knowledge extractor, gradient restorer and gradient integrator. In FedKNOW, each client has its private sequence of tasks. Upon receiving a new task $t_{m+1}$ in client $j$, the client needs to train for multiple iterations locally and then send back the trained model to the central server for global model aggregation.
Suppose the knowledge of $m$ previously learned tasks are retained ($m\geq 1$), FedKNOW trains the DNN model using $r$ aggregation rounds. Each round consists of two parts: local training with $v$ training iterations and global aggregation with the central server.}
\textcolor{black}{FedKNOW is designed with three objectives.}











\textcolor{black}{\textbf{Lightweight and scalable method}. In order to effectively training tasks on federated clients with a large number of tasks and clients, FedKNOW is a client-side method that exploits the limited resource of edge devices by extracting critical knowledge and integrating those of signature tasks.
The \textbf{knowledge extractor} retains each task $t_i$'s knowledge $W_i$, which corresponds to a set of the most important model weights. The \textbf{gradient restorer} transforms $W_i$ into gradient $g_i$ that represents the model update direction that maximizes task $t_i$'s accuracy. The above two components employ the weight-based pruning technique~\cite{han2016dsd,han2015learning} for two reasons: (1) it can filter and remove most of unimportant model weights and the retained weights in knowledge $W_i$ can precisely restore task $t_i$'s the gradient information; (2) the pruning phase is independent of the network architecture and completes quickly on edge devices. It is feasible to extend the above knowledge extraction and restoring process with structured pruning techniques such as L1-norm or L2-norm filter pruning~\cite{li2016pruning}
}


\textcolor{black}{\textbf{Catastrophic forgetting prevention}. At each training iteration, the \textbf{gradient integrator} takes task $t_{m+1}$'s original gradient $g_{m+1}$ and the $k$ gradients ($k \leq m$) of $t_{m+1}$'s most dissimilar previous tasks as inputs, and outputs an integrated gradient $g'$ that has an acute angle with all input gradients.
In a geometric term, this means updating the model using gradient $g'$ does not increase the loss (i.e. decreasing the accuracy) of the task represented by any input gradient~\cite{lopez2017gradient}. Among all previous tasks, the $k$ tasks that are most dissimilar with task $t_{m+1}$ are considered because the included angles between their gradients and $g_{m+1}$ are the largest. The integration process is solved as an optimization problem that minimizes the rotated angle between $g'$ and $g_{m+1}$ and completes in polynomial time.}

\textcolor{black}{\textbf{Negative knowledge transfer prevention}. Following the standard federated learning setting~\cite{li2019convergence}, the global model in the central server starts from a random client's model. At aggregation round $i$, client $j$ first uploads its local model weights to the server and obtains the updated global model after aggregation. After each global communication, FedKNOW fine tunes the model using one epoch of local samples. At each iteration, the \textbf{gradient integrator} takes the gradient $g^b_{m+1,i}$ before aggregation and $g^a_{m+1,i}$ after aggregation as inputs and outputs the integrated gradient $g^{'}_{m+1,i}$ that has acute angles to both input gradients. Hence using $g^{'}_{m+1,i}$ to update model can incorporate global information from other clients, while avoiding decreasing model accuracy in local data.
}



\subsection{Knowledge Extractor} \label{Section: Knowledge Extractor}

The knowledge extractor is designed with two purposes: the retained knowledge can keep most of the task's information and it can be quickly processed. The extraction process utilizes the weight-based pruning technique~\cite{han2016dsd,han2015learning} to remove most of the model weights whose absolute values are lower than a given threshold and retain the remaining ones (e.g. 10\%) as the signature task knowledge. For example, this pruning technique can removes over 90\% of the weights in VGG-16~\cite{VGG} and causes negligible accuracy losses.

Formally, let $W$ be the whole set of model weights for task $t_i$ ($i \geq 1$), its knowledge $W_i$ is defined as a proportion $\rho$ of weights with the largest values:
The knowledge extraction process has three steps: step 1 first trains the model until convergency; step 2 then selects a ratio $\rho$ (e.g. 10\%) of weights with the highest values to form $t_i$'s knowledge $W_i$; finally, step 3 fine-tunes the weights in $W_i$ while keeping other model weights unchanged.




\begin{equation}
\begin{split}
\label{TaskKnowledge}
\begin{aligned}
W_i = \{w_i\ |\ w_i \in W\  and\  w_i \geq \rho \}
\end{aligned}
\end{split}
\end{equation}
where $\rho$ is a quantile that decides the ratio of top-ranked weights that are extracted as task $t_i$'s knowledge.



\subsection{Gradient Restorer} \label{Section: Gradient Generator}


The knowledge restorer is designed for resource-constrained edge devices and it can produce a previous task $t_i$'s gradient without storing its training samples.
This component takes the retained knowledge \{$W_1$, $W_2$,...,$W_m$\} of $m$ tasks and the training samples ($X_{m+1}$,$Y_{m+1}$) of current task $t_{m+1}$ as inputs and outputs the gradients \{$g_1$, $g_2$,...,$g_m$\} of the $m$ tasks. Formally, let $loss()$ be the \textcolor{black}{cross-entropy loss function (i.e. log loss function)}, task $t_i$'s gradient is calculated as:
\begin{equation}
\begin{split}
\label{TaskKnowledgeGrad}
g_{i} = \nabla\ loss(f(W,X_{m+1}),f(W_i,X_{m+1}))\ (1 \leq i \leq m)
\end{split}
\end{equation}
where $f(W,X_{m+1})$ represents the predicted labels of the current task, $f(W_i,X_{m+1})$ represents the predicted labels according to $t_i$'s retained knowledge, and $\nabla$ is the gradient operator of these two labels. In contrast, task $t_{m+1}$'s gradient is calculated using its ground truth labels $Y_{m+1}$: $g_{m+1} = \nabla \ loss(f(W,X_{m+1}),Y_{m+1})$.

With the increased number of tasks (that is, $m$ is large), FedKNOW only selects the $k$ gradients that are most dissimilar with task $t_{m+1}$'s gradient. That is, the distances (e.g. Wasserstein distance) between these gradients and $g_{m+1}$ are the largest, hence these $k$ gradients' corresponding tasks are mostly influenced by the model updating using gradient $g_{m+1}$. In training, only the selected $k$ gradients are calculated to save computational costs.
\textcolor{black}{Note that in FedKNOW, parameters $\rho$ and $k$ are set according to hyperparameter search: a value is selected that it produces the highest model accuracy within certain memory or time constraint on edge devices. For the ratio $\rho$ of retained weights, the constraint is the memory footprint of these weights. For the number $k$ of gradients, the constraint is each task's computational time.}







\subsection{Gradient Integrator} \label{Section: Gradient Integrator}


The \emph{gradient integrator} is developed to find a rotated gradient $g'$ that decreases the loss of the current task $t_{m+1}$ without increasing the losses of its signature tasks. \textcolor{black}{These tasks are $t_{m+1}$'s $k$ most dissimilar tasks in preventing catastrophic forgetting, and they are tasks from other clients in preventing negative knowledge transfer.} This requires the included angle between $g'$ and any gradient $g_{i}$ of these tasks being an acute angle~\cite{lopez2017gradient}, because these gradients decide the updating directions of model weights.
If $t_{m+1}$'s original gradient $g_{m+1}$ does not meet the above requirement, the integrator aims to minimize the rotation angle between $g'$ and $g_{m+1}$, so as to maximize the learned knowledge of task $t_{m+1}$. Formally, let $G$=\{$g_1$ to $g_k$\} be the set of previous gradients, the integrator employs the quadratic programming~\cite{lopez2017gradient} to solve this optimization problem with polynomial time complexity:

\begin{equation}
\begin{split}
\label{IntegratedGrad}
\begin{aligned}
 \min_{g'}\ & \frac{1}{2}||g_{m+1},g'||^2_2 \\
 s.t. \ & Gg' \geq 0
\end{aligned}
\end{split}
\end{equation}
\textcolor{black}{where $Gg' =|G||g| \cos \theta \geq 0$ means the included angle $\theta$ between any gradient in $G$ and $g'$ is an acute angle}.
In Equation (\ref{IntegratedGrad}), $\frac{1}{2}||g_{m+1},g'||^2_2$=$\frac{1}{2}(g')^\top g' - g_{m+1}g' +\frac{1}{2}(g_{m+1})^\top g_{m+1}$, where $\frac{1}{2}(g_{m+1})^\top g_{m+1}$ is a constant and can be removed. Hence the gradient integrator solves the dual problem of Equation (\ref{IntegratedGrad}) as:
\begin{equation}
\begin{split}
\label{DualQuadraticProblem}
&\min_{v}\ \frac{1}{2}v^\top G G^\top v + g^\top G^\top v  \\
&s.t.\ v \geq 0
\end{split}
\end{equation}

That is, the gradient integrator solves the dual optimization programming in Equation (\ref{DualQuadraticProblem}) to find $v$ and calculates the integrated gradient as:
\begin{equation}
\begin{split}
\label{IntegratedGradient}
g' = G^\top v + g_{m+1}
\end{split}
\end{equation}


\subsection{Running Example} \label{Section: Running Example}

\begin{figure}[htp]
\centering
  \includegraphics[scale=0.51]{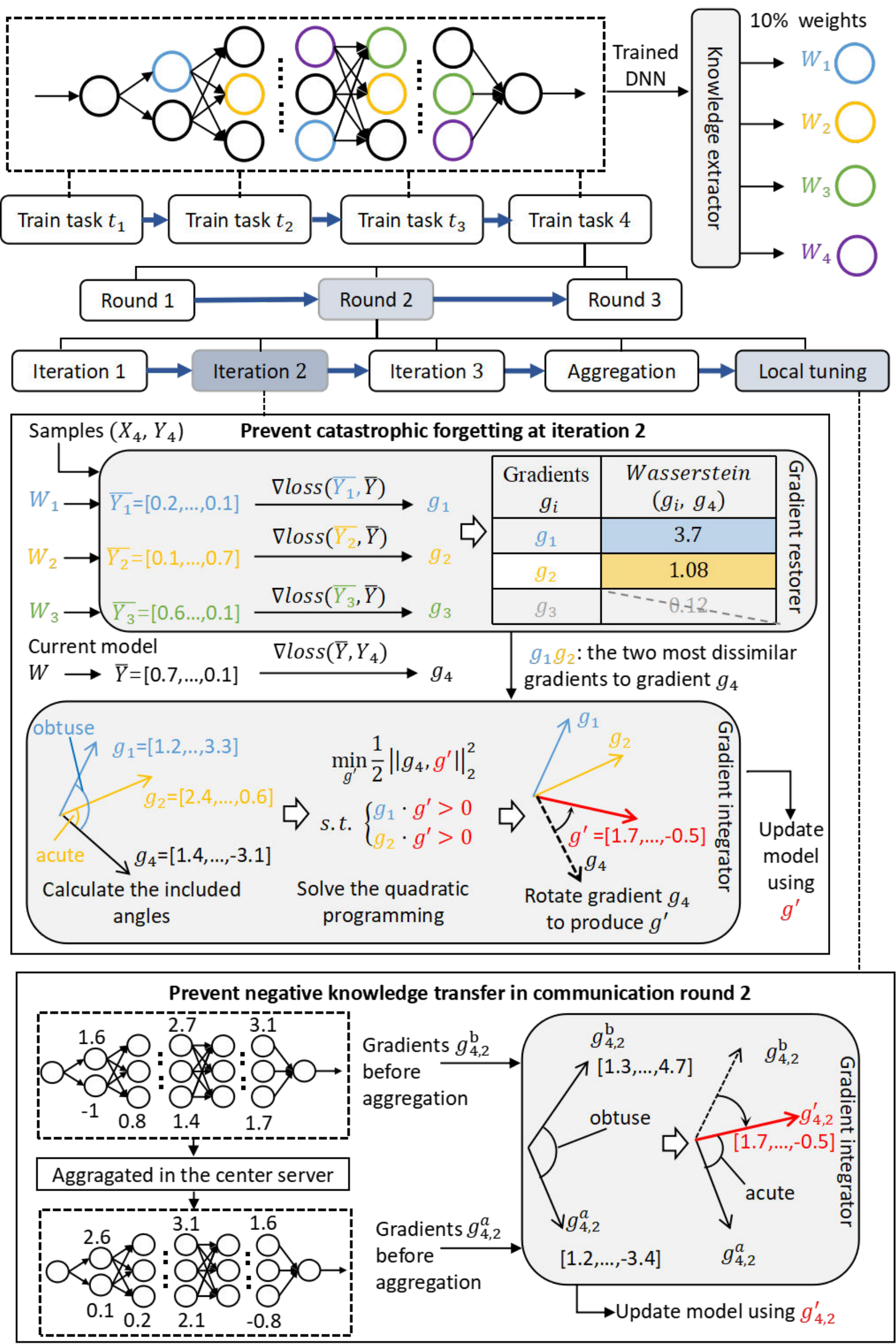}\\
  \caption{\textcolor{black}{Example model learning process with FedKNOW}}
  \label{Fig: ExampleFedKNOW}
\end{figure}

\textcolor{black}{Figure \ref{Fig: ExampleFedKNOW} shows FedKNOW's model training process when a new task $t_4$ arrives, and the whole process has three global aggregation rounds and each round has three local training iterations. After learning each task, the \emph{knowledge extractor} is applied to retain 10\% of model weights as this task's knowledge. This examples selects one iteration and one aggregation round to illustrates how FedKNOW works.
}

\textcolor{black}{
At iteration 2 of round 2, FedKNOW prevents \textbf{catastrophic forgetting} based on the retained knowledge ($W_1$ to $W_3$) of the three previously learned tasks ($t_1$ to $t_3$). The \emph{gradient restorer} first produces the labels of tasks $t_1$ to $t_3$ and uses these labels to compute gradients $g_1$ to $g_3$. It then calculates the Wasserstein distance between the gradient $g_i \ (1\leq i\leq 3)$ and $g_4$, and selects the two most dissimilar gradients $g_1$ and $g_2$.  Subsequently, the \emph{gradient integrator} computes the included angles between $g_4$ and two selected gradients ($g_1$, $g_2$), and finds that angle between $g_4$ and $g_1$ is obtuse. This means directly updating the model weights according to gradient $g_4$ will increase the loss function of task $t_1$ and degrade the accuracy of this task. The integrator thus solves the quadratic programming problem to find the minimal rotation angle to adjust $g_4$. Finally, the adjusted gradient $g¡¯$ is used in training.
}

\textcolor{black}{After completing three training iterations of round 2, the local mode is uploaded to the central server for global aggregation. FedKNOW then performs a fine tuning of the updated global model. At each tuning iteration, the \emph{gradient integrator} rotates each gradient before aggregation (e.g. $g_{4,2}^b$) such that it has an acute angle with the gradient after aggregation (e.g. $g_{4,2}^a$), and produces $g_{4,2}^{'}$ that is used to update the model. This updating direction incorporates the global information from other clients, while avoid their \textbf{negative knowledge transfer} to the local model before aggregation.
}

\section{Proof of Convergency in FedKNOW} \label{Section: Proof}

In this section, we prove the convergency of FedKNOW in the framework of federated learning and continual learning. 
Conceptually, the convergence means \emph{the model weights/parameters can achieve the global optimum ones over the training process}.
This work focuses on proving FedKNOW's convergency of model training in a client. For simplicity, we omit the index of the client in the following proof.

\textbf{Definition of convergency}. Let $W$ be the set of model weights in a client, $W^*$ be the optimal weights, and $W_{r}$ be weights at iteration $r$ ($r \geq 1$), and $f(\cdot)$ be the label prediction function of the model. During the iterative training process, the gap $H(r)$ at between these $W_{r}$ and $W^*$ is defined as:
\begin{equation}
\begin{split}
\label{Convergency}
H(r) =  \sum_{i=1}^r f(W_{r}) - \min_{W} \sum_{i=1}^r f(W)
\end{split}
\end{equation}

Given that $r$ is usually a large number and the \emph{training can converge} if $\frac{H(r)}{r}$ approaches 0. We convert Equation (\ref{Convergency}) as:
\begin{equation}
\begin{split}
\label{ConvergencyConvert} 
\lim_{r \rightarrow \infty}\frac{H(r)}{r}=  \lim_{r \rightarrow \infty} \mathbb{E}[f(W_r)] - f(W^*) = 0
\end{split}
\end{equation}
where $\mathbb{E}$ is the mathematical expectation. In convergency proof, we compute the upper bound of $\lim_{r \rightarrow \infty}\frac{H(r)}{r}$ and shows that it approaches 0 under some constraints. The proof is based on the three assumptions in existing work~\cite{zhang2012communication,stich2018local}
 
\textbf{Assumption 1.}\ \textit{The expected squared norm of stochastic gradients is uniformly bounded, i.e. $\mathbb{E}\Vert \nabla f(W_{r},\xi_{r})\Vert^2  \leq \lambda$, where $\xi_{r}$ is batch of training samples and $\lambda$ is a constant}

\textbf{Assumption 2.}\ \textit{The update of model parameters is bounded by a constant $D$: $\Vert W_r - W_{r+1}\Vert_2  \leq D$.}

In federated learning, suppose gradients follow assumptions 1 and 2, the upper bound of FedAvg~\cite{mcmahan2017communication} is given in assumption 3~\cite{li2019convergence}.

\textbf{Assumption 3.}\ 
\textit{In FedAvg, the training is bounded by:
\begin{equation}
\begin{split}
\label{FedAvgConvergence}
\mathbb{E}[f(W_{r})] - f(W^*) \leq \frac{\tau}{\gamma + r -1} ( \frac{2B}{\mu} +\frac{\mu \gamma}{2}\mathbb{E}||W_{r} - W^*||^2 ) \\
\end{split}
\end{equation}
where $B = \sum^N_{i=1} p_i^2 \sigma_i^2 + 6L\Omega + 8(r-1)^2\lambda^2$, $L$, $\mu$ are constant, $\sigma_i$ is the upper bound of gradient $g_r$'s variance,  $\lambda$ is the upper bound of $(g_r)^2$, $\tau = \frac{L}{\mu}$, $\gamma = max\{8\tau,r\}$, $p_i$ denotes the weights of client $i$, and $\Omega = f^*(W) - \sum_{i=1}^n p_if_i(W^*)$ denotes training data's degree of severity in terms of Non-IID. \
}
 
\textbf{Proof steps}. In FedKNOW, suppose $W_{r}= W_{r}^G \cup W_{r}^L$ consists of global weights $W_{r}^G$ and local weights $W_{r}^L$,  At iteration $r$, let $\eta^G_r$ and $\eta^L_r$ be the learning rates used in training weights $W_{r}^G$ and $W_{r}^L$, respectively. The proof of FedKNOW's convergency has three steps: \textbf{Lemma 1} proves the upper bound of training $W^G$; \textbf{Lemma 2} proves the upper bound of training $W^L$; and finally \textbf{Theorem 1} proves the convergence of training the whole model $W$ under the constraints of two learning rates $\eta^G_r$ and $\eta^L_r$.



\subsection{Upper bound of Training Local Weights $W^L$} \label{Section: Local}

\textbf{Lemma 1.}\ \textit{Let $W^L_{r}$ be the client's local weights at iteration $r$ and $W^{L^*}$ be the optimal local weights, the training of $W^L_{r}$ is bounded by: }

\begin{equation}
\begin{split}
\label{LocalProof}
\mathbb{E}[{f(W^L_{r})}] - f(W^{L^*}) \leq \frac{D^2}{2\eta^L_r r} + \frac{\lambda^2 \eta^L_j}{2}
\end{split}
\end{equation}

\textbf{Proof}.  Let $g_r=\nabla f(W^L_r, \xi_r)$ be the gradient at iteration $r$. According to Equation (\ref{Convergency}), the gap $H(r)$ between local weights $W_r^L$ and $W^{L^*}$ is:


\begin{equation}
\begin{split}
\label{Statistic}
H(r) & = \sum^r_{i=1}f(W^L_i) -\min_{W^L} \sum_{i=1}^r f(W^L_i) \\
& = \sum^r_{i=1} [f(W^L_i) - f(W^{L^*})]
\end{split}
\end{equation}


Suppose that $f(\cdot)$ is convex, we have:

\begin{equation}
\begin{split}
\label{Convex}
f(W^{L^*}) \geq f(W^L_r) + \langle g_r,W^L_r - W^{L^*} \rangle
\end{split}
\end{equation}

In model updating, $W^L_r$ satisfies:

\begin{equation}
\begin{split}
\label{GradExchange}
& W^L_{r+1} = W^L_{r} - \eta^L_r g_r \\
\longrightarrow\quad & W^L_{r+1} -  W^{L^*} = W^L_{r} -W^{L^*} - \eta^L_r g_r \\
\longrightarrow\quad & ||W^L_{r+1} -  W^{L^*}||^2_2 = ||W^L_{r} -W^{L^*} - \eta^L_r g_r||^2_2 \\
\longrightarrow\quad & \langle g_r,W^L_r - W^{L^*} \rangle = \frac{1}{2\eta^L_r} ( ||W^L_{r} -W^{L^*}||^2_2-  \\ &\quad ||W^L_{r+1} -W^{L^*}||^2_2) + \frac{\eta^L_r}{2}||g_r||^2_2
\end{split}
\end{equation}

By combining Equations (\ref{Statistic}), (\ref{Convex}), and (\ref{GradExchange}) we have:

\begin{equation}
\begin{split}
H(r)
&  \leq \sum_{j=1}^r \frac{1}{2\eta^L_j} ( ||W^L_{j} -W^{L^*}||^2_2- ||W^L_{j+1} -W^{L^*}||^2_2 )\\  &\quad +  \sum_{j=1}^r \frac{\eta^L_j}{2}||g_j||^2_2
\end{split}
\end{equation}

We further lower the upper bound $H(r)$ of local weights $W^L$ based on the Assumptions 1 and 2 and scale $H(r)$ as:

\begin{equation}
\begin{split}
\label{WLConverge}
& H(r) \leq D^2\frac{1}{2\eta^L_j} + \frac{\lambda^2}{2}\sum_{j=1}^r \eta^L_j \\
\longrightarrow\quad & \mathbb{E}[{f(W^L_r)}] - f(W^{L^*}) \leq \frac{D^2}{2\eta^L_r r} + \frac{\lambda^2 \eta^L_j}{2} 
\end{split}
\end{equation}


\subsection{Upper bound of Training Global Weights $W^G$} \label{Section: Global}

\textbf{Lemma 2.}\ \textit{Let $W^G_r$ be the client's global weights at iteration $r$ and $W^{G^*}$ be the optimal global weights, the training of $W^G_{r}$ is bounded by: }
\begin{equation}
\begin{split}
\label{GlobalCovergeThrorem}
&\mathbb{E}[f(W^G_{r})] - f(W^{G^*}) \leq \\
&\frac{\tau}{\gamma + r -1} ( \frac{2B}{\mu} +\frac{\mu \gamma}{2}\mathbb{E}||W^G_{r} - W^{G^*}||^2 ) \\
\end{split}
\end{equation}

\textit{where $B = \sum^N_{i=1} p_i^2 \sigma_i^2 + 6L\Omega + 8(r-1)^2(g')^2$.}

\textbf{Proof}. FedKNOW employs FedAvg~\cite{li2019convergence} as the global parameter aggregation algorithm. Hence if its gradient $g'$ follows the Assumptions 1 and 2, its training of global weights can be bounded by Equation (\ref{GlobalCovergeThrorem}) according to Assumption 3. We now proves the boundedness of $g'$.

In FedKNOW, $g'$ is an integrated gradient of $k$ previous gradients ($G$=\{$g_1$ to $g_k$\}) and the gradient $g_r$ calculated using the current task's training samples. That is, $g' = G^\top v + g_{r}$ ($v \geq 0$) according to Equation (\ref{IntegratedGradient}). According to Assumption 1, all gradients in $G$ are bounded and $g_{r}$ is a constant, we have $||g'||^2$ in Equation (\ref{GlobalCovergeThrorem}) is bounded:
\begin{equation}
\begin{split}
\label{BoundedGradient}
    & ||g'||^2 = max (||G^\top v +  g_{r}||^2) \\
 &= max( g_{r}^\top  g_{r}  + 2v^\top G  g_{r}+ 1)
\end{split}
\end{equation}


\subsection{Convergence of Overall Model} \label{Section: Overall}

\textbf{Theorem 1.}\ \textit{In a client, FedKNOW can converge under two constraints: (1) its local weights's learning rate $\eta^L$  decreases at the rate of $\mathcal{O}(r^{-\frac{1}{2}})$; and (2) its global weights' learning rate $\eta^G \leq \frac{2}{\mu (\gamma + r)}$ and it decreases at the rate of $\mathcal{O}(r^{-1})$:
\begin{equation}
\begin{split}
\label{OvrallConvergency}
\lim_{r \rightarrow \infty} \mathbb{E}[f(W^L_{r} \cup W^G_{r})] - f(W^*) = 0
\end{split}
\end{equation}
}

\textbf{Proof}. Let $W^*=W^{L^*} \cup W^{G^*}$, we convert Equation (\ref{OvrallConvergency}) as:

\begin{equation}
\begin{split}
\label{OvrallConvergencyExtend}
    & \lim_{r \rightarrow \infty} \mathbb{E}[f(W^L_{r} \cup W^G_{r})] - f(W^{L^*} \cup W^{G^*}) = 0 \\
    \longrightarrow & \lim_{r \rightarrow \infty} (\mathbb{E}[f(W^L_{r}]- f(W^{L^*}) \cup (\mathbb{E}[f(W^G_{r}]- f(W^{G^*})= 0
\end{split}
\end{equation}

According to Lemma 1, $\lim_{r \rightarrow \infty} \mathbb{E}[f(W^L_{r})]-f(W^{L^*})$ is bounded (Equation (\ref{LocalProof})) and this bound approaches to 0 if the learning rate $\eta^L$ decreases at the rate of $\mathcal{O}(r^{-\frac{1}{2}})$~\cite{lecun2015deep,zinkevich2003online}.
Similarity, Lemma 2 states that $\lim_{r \rightarrow \infty} \mathbb{E}[f(W^G_{r})]-f(W^{G^*})$ is bounded (Equation (\ref{GlobalCovergeThrorem})) and this bound approaches to 0 if learning rate $\eta_r^G \leq \frac{2}{\mu (\gamma + r)}$ and decreases at the rate of $\mathcal{O}(r^{-1})$~\cite{li2019convergence}.
$\blacksquare$


\section{Evaluation} \label{Section: Evaluation}

In this section, we evaluate the full implementation of FedKNOW on top of PyTorch~\cite{PyTorch} on exhaustive experimental scenarios against a wide set of data benchmarks and DNNs.

\subsection{Experimental Settings} \label{Sec: Experimental Settings}

\textbf{Testbed}. We choose four types of heterogeneous edge platforms imposing different architectures to showcase FedKNOW's cross-platform nature when it comes to hardware: Jetson TX2 has 256-core NVIDIA Pascal GPU and 8 GB memory; Jetson Nano has NVIDIA Maxwell architecture with 128 NVIDIA CUDA cores and 4GB memory; Jetson Xavier NX has 384-core NVIDIA Volta GPU with 48 Tensor Cores and 16 GB memory; and Jetson AGX has 512-core Volta GPU with Tensor Cores and 32 GB memory.
All Jetson platforms run Ubuntu 18.04.5 LTS and support DNNs in PyTorch 1.9.0 (Python 3.6.9).

\textbf{Datasets and DNN models}. We select five representative federated and continual datasets to evaluate FedKNOW.
\textcolor{black}{In these datasets, a \emph{task} refers to an image classification task for a given set of objects.
Following the setting of typical continual learning methods~\cite{rebuffi2017icarl}, the training/test points in each dataset are equally splited into each task and each class.}

\begin{itemize}[noitemsep,nolistsep]
\item \emph{Cifar100~\cite{krizhevsky2009learning}} and \emph{FC100~\cite{oreshkin2018tadam}} datasets both have 50k data points (training samples) from 100 classes and 10k testing points (100 ones per class). In continual learning, these data points belong to 10 tasks and each task has 10 classes.
\item \emph{CORe50~\cite{lomonaco2017core50}} dataset has 165k data points from 550 classes and 55k testing points (100 ones per class). These data points belong to 11 tasks and each task has 50 classes.
\item \emph{MiniImageNet~\cite{vinyals2016matching}} dataset has 50k data points from 100 classes and 10k testing points (100 ones per class). These data points belong to 10 tasks and each task has 10 classes.
\item \emph{TinyImageNet~\cite{le2015tiny}} dataset has 100k data points from 200 classes and 10k testing points (50 ones per class). These data points belong to 20 tasks and each task has 10 classes.
\end{itemize}



In evaluation, the first three datasets are trained with a 6-layer CNN model~\cite{jung2020continual} and the last two datasets are trained with the ResNet-18 model~\cite{he2016deep}. To evaluate FedKNOW's generalization capability on different network architectures, we also test it using 8 state-of-the-art DNNs with different depths, widths, multi-path, feature map exploitation mechanisms~\cite{khan2020survey}.


\textcolor{black}{\textbf{Task and dataset assignment in federated setting}. We set the distributions of tasks and datasets following the setting of FedRep~\cite{collins2021exploiting}. Each client has all tasks of a dataset and its distinct task sequence.
To guarantees the data heterogeneously (non-IID) among different clients, we randomly allocates 2 to 5 of each task's classes to each client.
For each class, we randomly selects 5\% to 10\% of the training samples in allocation.
}




\textbf{Compared baselines}. We implement and compare our approach with 11 state-of-the art techniques that can be divided into three categories£º
\begin{itemize}[noitemsep,nolistsep]
\item Six \emph{continual learning} methods: (1) gradient episodic memory (GEM) for continual learning~\cite{lopez2017gradient,chaudhry2018efficient} calculates previous gradients and uses the included angle between them and the current gradient in model training.
(2) Balanced Continual Learning (BCN)~\cite{raghavan2021formalizing} retains the previous training samples and uses them to maximize the data distribution among different tasks and minimize the model training errors.
(3) Contrastive Continual Learning($Co^2L$)~\cite{cha2021co2l} focuses on feature transfer and maintains contrastive learned representations to mitigate catastrophic forgetting.
(4) Elastic Weight Consolidation(EWC)~\cite{kirkpatrick2017overcoming} initially proposes the idea of regularization. It uses the Fisher information matrix to calculate the changes of model weights in different tasks and avoid drastic changes in weights.
(5) Memory aware synapses(MAS)~\cite{aljundi2018memory} improves this approach by estimating each weight's importance according to the output's sensitivity to this weight.
And (6) adaptive Group Sparsity Based Continual Learning (AGS-CL)\cite{jung2020continual} applies different learning strategies for different tasks.

\item Three \emph{federated learning} methods:
(1) FedAvg~\cite{mcmahan2017communication} is a typical approach that calculates each client's weight factor according to its number of training samples and uses these factors to aggregate the models of all clients.
(2) APFL~\cite{deng2020adaptive} dynamically changes the ratio of global and local models in training.
And (3) FedRep~\cite{collins2021exploiting} divides a model into presentation layers and head layers, and only communicates presentation layers in federated learning, while adaptively training model weights in each client.

\item  Two \emph{Federated continual learning} methods:
(1) federate learning with continual local training (FLCN) maintains some training samples in the server and uses them in global model updating to avoid catastrophic forgetting~\cite{yao2020continual}.
(2) FedWEIT~\cite{yoon2021federated} is the latest technique that uses masks to divide model weights into base ones and adaptive ones, and maintains the adaptive weights of all clients and tasks as the previous knowledge. In each client, it obtains all clients' adaptive parameters and trains them together with the new task's weights based on the regularization method.
\end{itemize}



\textbf{Evaluation Metrics}. We consider both model accuracy and training time (hour) in evaluation.
The accuracy metric is the top-1 accuracy on test data points: the top predicted class (the one with the largest probability) is the same as the actual class label. In a continual learning scenario, the reported accuracy of task $t_m$ is the average accuracy of all $m$ learned tasks.





\subsection{Comparative Evaluations under Different FCL Scenarios} \label{Sec: Effectiveness}

This section's evaluation compares \emph{model accuracy} and \emph{training time} between FedKNOW and 11 baseline techniques. \textcolor{black}{To make our comparisons fair and avoid leakage of test data, we employ the prevalent benchmarking method~\cite{gulrajani2020search} that searches hyperparameter using an additional test dataset (that is, SVHN~\cite{netzer2011reading} with two tasks and each class has 5 classes). For each dataset, this method searches the optimal hyperparameters that produce the highest accuracy on the SVHN dataset.} The model training settings include two parts:
\begin{itemize}[noitemsep,nolistsep]
\item \textbf{Common training settings for all techniques}.  In comparison, the model is trained using the same initial weights, training samples, hyperparameters, and a cluster of 20 heterogeneous edge clients, including 2 Jetson AGX, 2 Jetson TX2, 8 Jetson Xavier NX, and 8 Jetson Nano platforms.
    \textcolor{black}{In hyperparameter search, the search scopes of aggregation rounds and training iterations are 5 to 15 and 5 to 150, respectively. The search scopes of learning rates and decreases rates are \{0.0005,0.0008,0.001,0.005\} and \{1e-6,1e-5,1e-4\}. The scopes of both hyperparameters satisfy the condition of convergency in Section \ref{Section: Proof}.
    For Cifar100, FC100, CORe50, MinyImageNet, and TinyImageNet workloads, the numbers of global aggregation rounds are set to 15, 15, 15, 10, and 5, respectively. Each round consists of 25 local local training iterations (i.e. 5 epoches). In these five workloads, the learning rates are set to 0.001, 0.001, 0.001, 0.0008, and 0.0008, and their decrease rates are set to 1e-4, 1e-4, 1e-4, 1e-5, and 1e-5, respectively.
    }

\item \textbf{Specific settings for some techniques}. \textcolor{black}{In hyperparameter search of each baseline method, we set the lower and upper bounds of search space as 1/2 and 2 of the parameter value in its original evaluation.} In memory-based continual learning methods(GEM, BCN, $Co^2L$), 10\% of training samples are retained to avoid catastrophic forgetting. In regularization-based continual learning methods, the regularization hyper-parameters are 40000 and 100 for EWC and MAS. In FLCN, we select 10\% of training samples randomly to the server for updating the regularization parameters.
    \textcolor{black}{
    In FedKNOW, the search space of ratio $\rho$ of retained weights and number $k$ of selected gradients in integration are $\{5\%,10\%,20\%\}$, and $\{5,10,20\}$, respectively.
    Ratio $\rho$ is set to 10\% because 20\% of weights exceed the memory constraint of 4 GB and 10\% produces a better accuracy than that of 5\%.
    Gradient number $k$ is set to 10 because it produces the highest accuracy within the time constraint (that is, each task's computational time is smaller than 20 minutes).
    }

\end{itemize}

\begin{figure*} [htp]
\centering
  \includegraphics[scale=0.4]{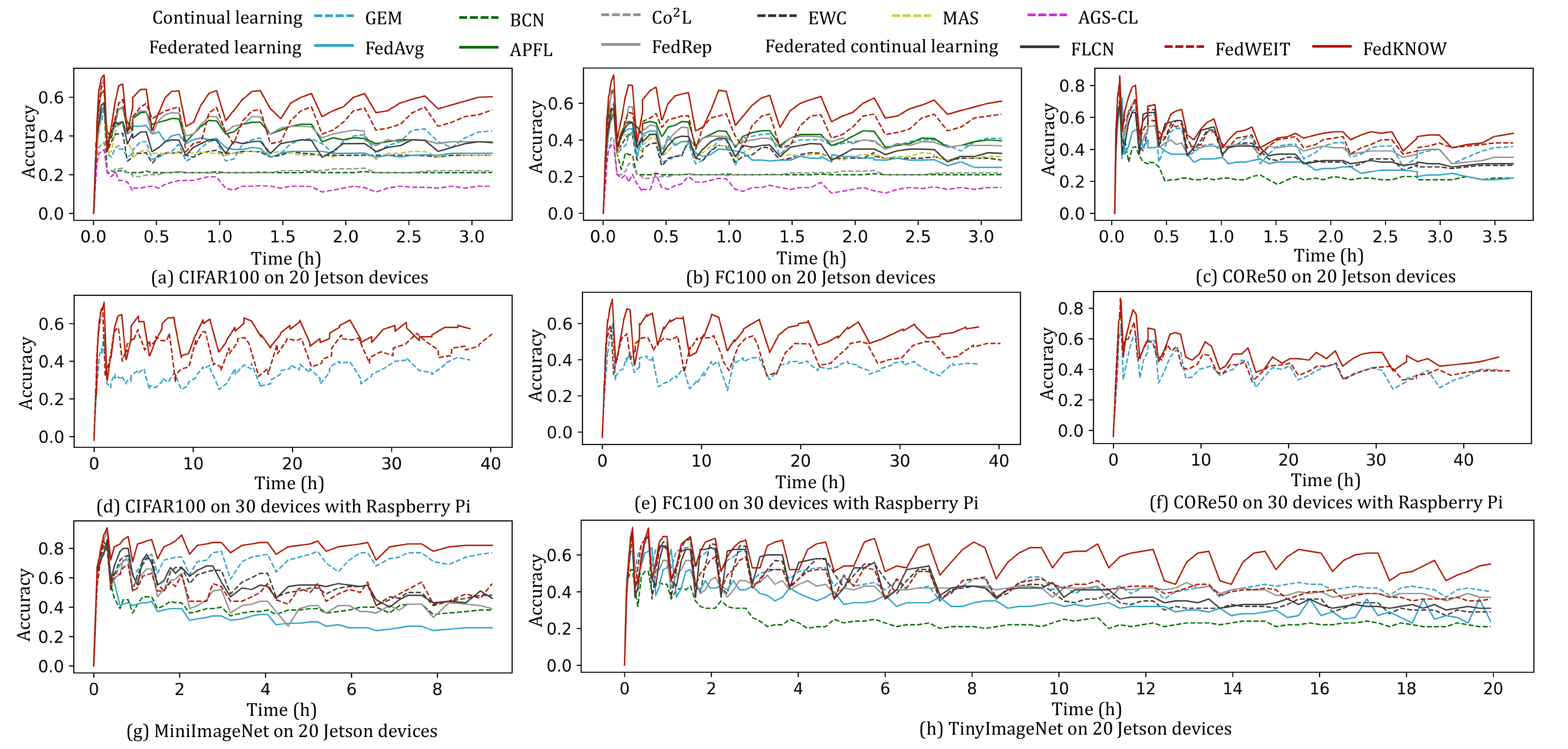}\\
  \caption{\textcolor{black}{Comparison of model accuracy and training time between FedKNOW and 11 baseline methods}}
  \label{Fig: ComparisonBasic}
\end{figure*}

\textbf{Comparison results}. Figure \ref{Fig: ComparisonBasic} displays the comparison results of 12 techniques and we have three key observations.

\textcolor{black}{\emph{Impact of catastrophic forgetting}. Three federated learning baselines take less time to converge because these methods donot consider previous task information in model training, hence their model accuracies are lower than most of the other techniques due to catastrophic forgetting.
This also explains the results that when the number of task increases, the accuracies of all techniques decrease. FedKNOW suffers least from the accuracy depredation because when learning a new task, it integrates its knowledge with the seen tasks that are most dissimilar from the current task model. In contrast, FedWEIT uses the maintained knowledge of all tasks (stored at the server) and may lower the influence of important tasks.
}

\textcolor{black}{
\emph{Impact of negative knowledge transfer}. In a federated learning environment, the non-IID datasets in different clients also considerably influence model accuracy. The five continual learning baselines well address catastrophic forgetting, but suffer from negative knowledge transfer from other clients. For example, AGS-CL's loss function considers the changes in model weights. Hence the large changes in global model weights cause non-convergence in CORe50, MiniImageNet, and TinyImageNet datasets, and this observation is no observed in Figures \ref{Fig: ComparisonBasic}(c), (g), and (h). In addition, FedWEIT has higher accuracies than other baseline methods in the first three datasets.
However, its parameter decomposition strategy may harm the functionalities of some particular layers (downsample in ResNet) and thus its accuracies are lower in this DNN model (Figures \ref{Fig: ComparisonBasic} (g) and (h)). In contrast, FedKNOW achieves the highest accuracies in all settings thanks to its reliable gradient integration mechanism.
}


\textcolor{black}{
\emph{Impact of heterogeneous edge devices}. We extend the above evaluation by adding 10 CPU-based devices (Raspberry Pi 4B) to the cluster with 20 Jetson devices. The Raspberry Pi devices consist of one with 2 GB memory, five ones with 4 GB memory, and four ones with 8 GB memory. Using the Cifar100, FC100, and CORe50 datasets, this evaluation compares the three techniques (GEM, FedWEIT, and FedKNOW) that produce the highest accuracies among all techniques. Figures \ref{Fig: ComparisonBasic}(d), (e), and (f) show that: (i) training in resource-limited Raspberry Pi devices considerably delays the training time of all techniques by an average of 12 times. In particular, FedWEIT has the largest increase in training time because its global knowledge becomes larger with more tasks and clients. For example, FedWEIT's training time of the last task is 1 hour longer than the other two techniques. (ii) Resource heterogeneity decreases accuracies for all three techniques by 3\% to 5\%. The results show FedKNOW still achieves the highest accuracies because it is lightweight and integrates task knowledge locally. In contrast, FedWEIT requires each client using the heavyweight global knowledge, which makes the Raspberry Pi of 2 GB memory out of memory after learning 7 tasks and cannot participate in the following federated learning.
}




\textcolor{black}{Table \ref{tab: AccuracySummary} summarizes the percentages of increase in the average accuracy, when comparing the accuracy of FedKNOW against the average accuracy of all 11 baselines techniques across 5 different datasets}. For each dataset, the increased accuracy of each task is reported.
We can see that when the task number increases, the percentage accuracy improvement increases from 10.21\% to 98.72\%.
Overall, when considering all evaluation cases, our approach improves the accuracy by an average of 77.35\%, 33.26\%, and 31.27\% compared to continual learning, federated learning, and FCL baselines, respectively.


\begin{table}[]
    \caption{A summary of average percentage accuracy improvement}
	\renewcommand \arraystretch{1.3}
	\centering
	\resizebox{\linewidth}{!}{%
		\begin{tabular}{|c|c|c|c|c|ccc|}
			\hline
			& \textbf{CIFAR100} & \textbf{FC100} & \textbf{Corn50} & \textbf{\begin{tabular}[c]{@{}c@{}}Mini\\ Imagenet\end{tabular}} & \multicolumn{3}{c|}{\textbf{\begin{tabular}[c]{@{}c@{}}Tiny\\ Imagenet\end{tabular}}} \\ \hline
			\textbf{Task1}  & 36.52\%           & 36.16\%        & 17.31\%         & 10.21\%                                                          & \multicolumn{1}{c|}{16.00\%}    & \multicolumn{1}{c|}{\textbf{Task11}}    & 64.00\%   \\ \hline
			\textbf{Task2}  & 74.74\%           & 63.62\%        & 38.80\%         & 20.39\%                                                          & \multicolumn{1}{c|}{18.31\%}    & \multicolumn{1}{c|}{\textbf{Task12}}    & 67.98\%   \\ \hline
			\textbf{Task3}  & 82.58\%           & 74.89\%        & 32.54\%         & 37.04\%                                                          & \multicolumn{1}{c|}{20.62\%}    & \multicolumn{1}{c|}{\textbf{Task13}}    & 73.00\%   \\ \hline
			\textbf{Task4}  & 84.84\%           & 79.22\%        & 46.21\%         & 70.30\%                                                          & \multicolumn{1}{c|}{28.22\%}    & \multicolumn{1}{c|}{\textbf{Task14}}    & 80.27\%   \\ \hline
			\textbf{Task5}  & 88.69\%           & 86.83\%        & 45.39\%         & 68.12\%                                                          & \multicolumn{1}{c|}{33.00\%}    & \multicolumn{1}{c|}{\textbf{Task15}}    & 84.00\%   \\ \hline
			\textbf{Task6}  & 94.87\%           & 86.15\%        & 37.27\%         & 70.57\%                                                          & \multicolumn{1}{c|}{35.09\%}    & \multicolumn{1}{c|}{\textbf{Task16}}    & 93.50\%   \\ \hline
			\textbf{Task7}  & 92.40\%           & 88.81\%        & 37.06\%         & 65.61\%                                                          & \multicolumn{1}{c|}{45.00\%}    & \multicolumn{1}{c|}{\textbf{Task17}}    & 95.00\%   \\ \hline
			\textbf{Task8}  & 95.52\%           & 85.50\%        & 51.84\%         & 67.98\%                                                          & \multicolumn{1}{c|}{54.46\%}    & \multicolumn{1}{c|}{\textbf{Task18}}    & 97.78\%   \\ \hline
			\textbf{Task9}  & 98.72\%           & 87.57\%        & 56.65\%         & 70.34\%                                                          & \multicolumn{1}{c|}{52.00\%}    & \multicolumn{1}{c|}{\textbf{Task19}}    & 91.00\%   \\ \hline
			\textbf{Task10} & 97.75\%           & 90.49\%        & 54.72\%         & 72.18\%                                                          & \multicolumn{1}{c|}{61.49\%}    & \multicolumn{1}{c|}{\textbf{Task20}}    & 87.57\%   \\ \hline
			\textbf{Task11} &                   &                & 63.22\%         &                                                                  & \multicolumn{1}{c|}{}           & \multicolumn{1}{c|}{}                   &           \\ \hline
		\end{tabular}%
	}
\label{tab: AccuracySummary}
\end{table}

\subsection{Evaluation of Communication Cost} \label{Sec: Communications}

Following the evaluation settings of the previous section, this section's evaluation focuses on communication cost in model training. We compare our approach with FedWEIT because in this method, each client needs to obtain the retained adaptive weights of all other clients before learning a new task. Although these weights bring higher accuracies, they also incur large communication traffic that increases with the number of clients.
In contrast, both our approach and other baseline methods employs the standard FedAvg method in federated learning and have the same communication cost.

\begin{figure}[htp]
\centering
  \includegraphics[scale=0.47]{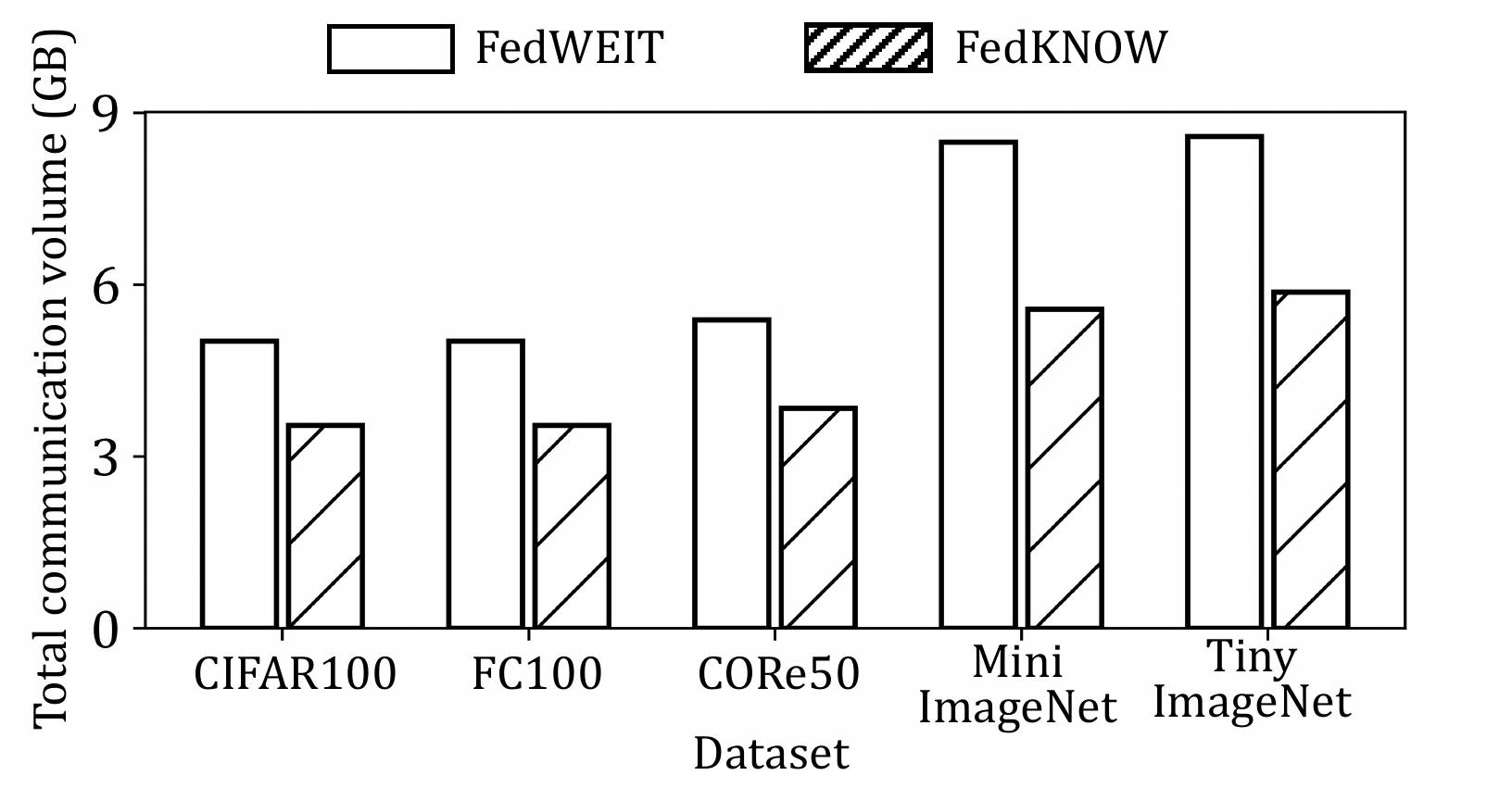}\\
  \caption{\textcolor{black}{Comparison of communication time under different workloads}}
  \label{Fig: CommunicationTime}
\end{figure}

\textbf{Evaluation of different workloads}. In federated learning, the communication cost among clients and the central server considerably impacts the model training performance (communication time takes about 10\% to 30\% of model training time). As shown in \ref{Fig: CommunicationTime}, FedKNOW takes much less communication cost when performing the same model training task. This is because our approach employs a distributed knowledge retaining mechanism that each client only uses its own knowledge to retain previous tasks.
By contrast, FedWEIT applies a centralized mechanism that aggregates all clients' adaptive weights in the server and uses these as the knowledge. This means each time a client learns a new task, it needs to send its latest model weights to the server and all other clients need to obtain these weights from the server. We note that in FedWEIT, a client's own adaptive weights cannot represent its previous tasks. This is these parameters are generated using regularization techniques and in a client, different tasks' adaptive weights have small differences. Hence FedWEIT needs to use other clients' adaptive weights to increase the model generalization in continual learning.
Overall, our approach reduces communication cost by 34.28\%.


\begin{figure}[htp]
\centering
  \includegraphics[scale=0.49]{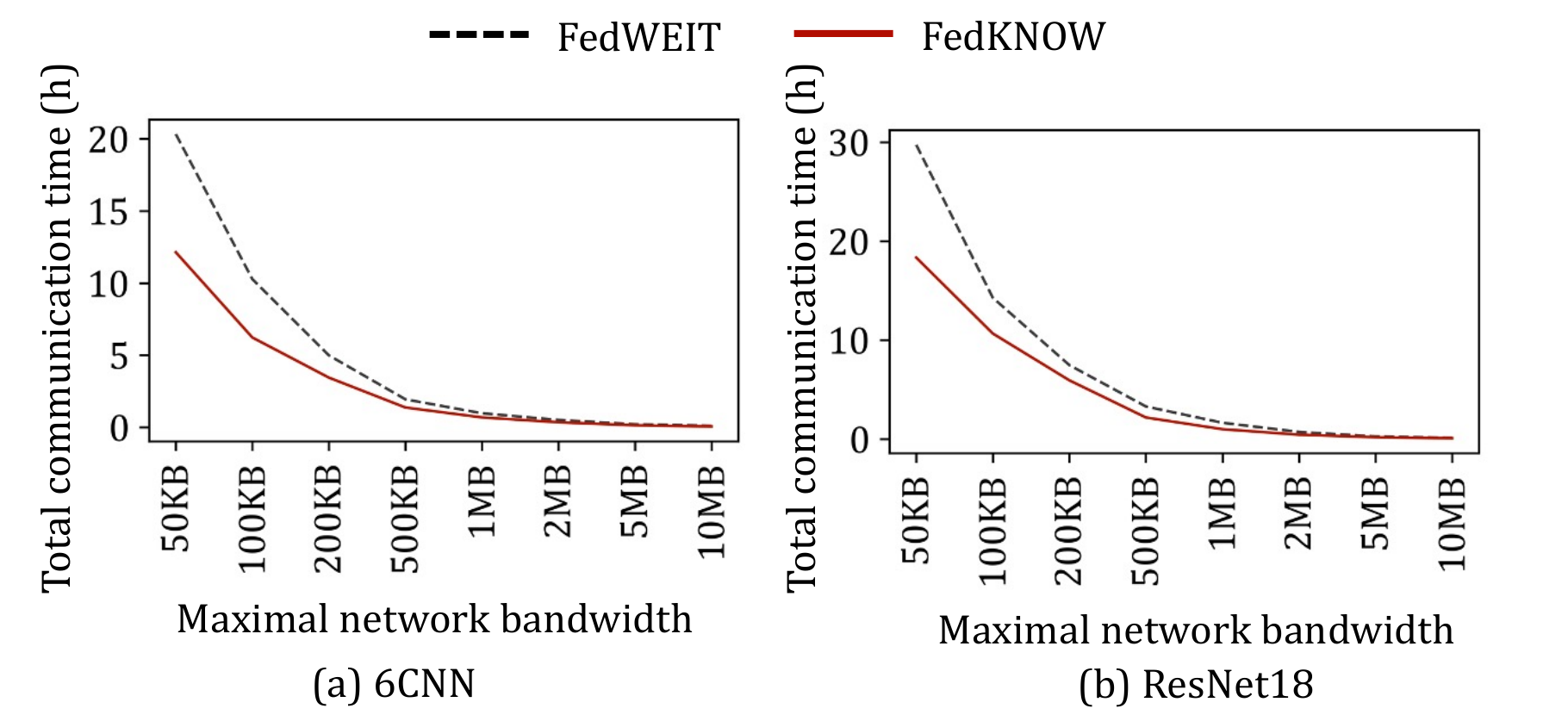}\\
  \caption{\textcolor{black}{Comparison of communication time under different network bandwidths}}
  \label{Fig: CommunicationBandwith}
\end{figure}

\textbf{Evaluation under different network bandwidths}. In a distributed edge computing environment, network bandwidth is a key factor that influences communication time. In the previous evaluation, the network bandwidth limitation is 1 MB/second. We extended this evaluation be test 8 different network bandwidths, ranging from 50 KB per second to 10 MB per second, in each client.
Figure \ref{Fig: CommunicationBandwith} shows the communication time of two DNN models under different bandwidths.
We can see that our approach consistently takes less communication time than FedWEIT.
As expected, the communication time becomes longer when the network bandwidth decreases and our approach can save more communication time under tenser network conditions (up to 10 hours when the network bandwidth is 50 KB per second).


\subsection{Discussion of Task and Client Numbers} \label{Sec: Communications}

In this section, we extended the previous evaluations to discuss two key factors in federated and continual learning scenarios: the number of task and the number of client. By taking FedWEIT, GEM, and FedKNOW as examples, our discussion demonstrates how these two factors effect the learning effectiveness. This is because FedWEIT and GEM perform best in all baselines: FedWEIT is developed specifically for the continual setting and GEM stores the past samples directly which sacrifices the efficiency to mitigate catastrophic forgetting.
In evaluation, we report two metrics: \emph{average accuracy} and \emph{average forgetting rate} in all learned tasks.
Suppose $m$ tasks are learned, the forgetting rate of the $k$th task ($1\leq k \leq m$) is defined as: the difference between $t_k$'s accuracies after learning $k$ and $m$ tasks, divided by the former accuracy. We report the forgetting rate after learning a new task and its value ranges between 0 and 1.

\begin{figure}[htp]
\centering
  \includegraphics[scale=0.48]{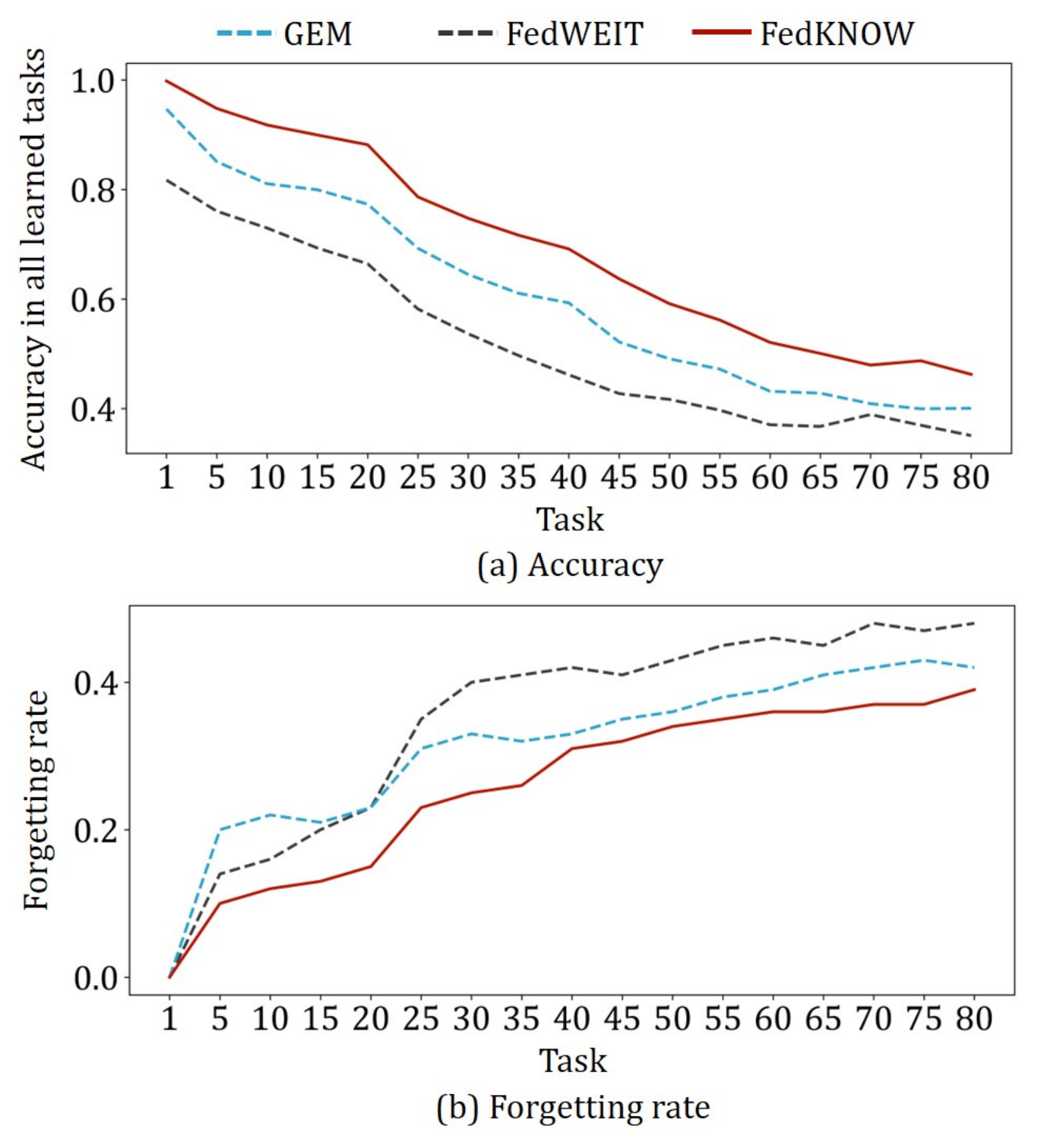}\\
  \caption{\textcolor{black}{Discussion of model accuracies under different numbers of tasks}}
  \label{Fig: DiscussTask}
\end{figure}

\textbf{Number of tasks}. In this evaluation, we combine the tasks in MiniImageNet, Cifar100, and TinyImage workloads, and construct a dataset with 80 tasks. We still use FedRep's method~\cite{collins2021exploiting} to distribute this dataset to 20 clients to guarantee data heterogeneity, and use ResNet-18 to learn these tasks. Figure \ref{Fig: DiscussTask} shows the fluctuations of accuracies and forgetting rates when the number of tasks increases from 1 to 80. The comparison results show that:
(i) FedKNOW consistently provides higher latencies. This result verifies that our approach maintains the most of previous tasks' knowledge among three methods.
(ii) When the job number increases, the model accuracies have apparent decreases in all methods. This result can be explained by these methods' forgetting rates: ResNet-18 has a limited generalization capacity and hence its forgetting rate continuously increases when more task knowledge is learned.

\begin{figure}[htp]
\centering
  \includegraphics[scale=0.48]{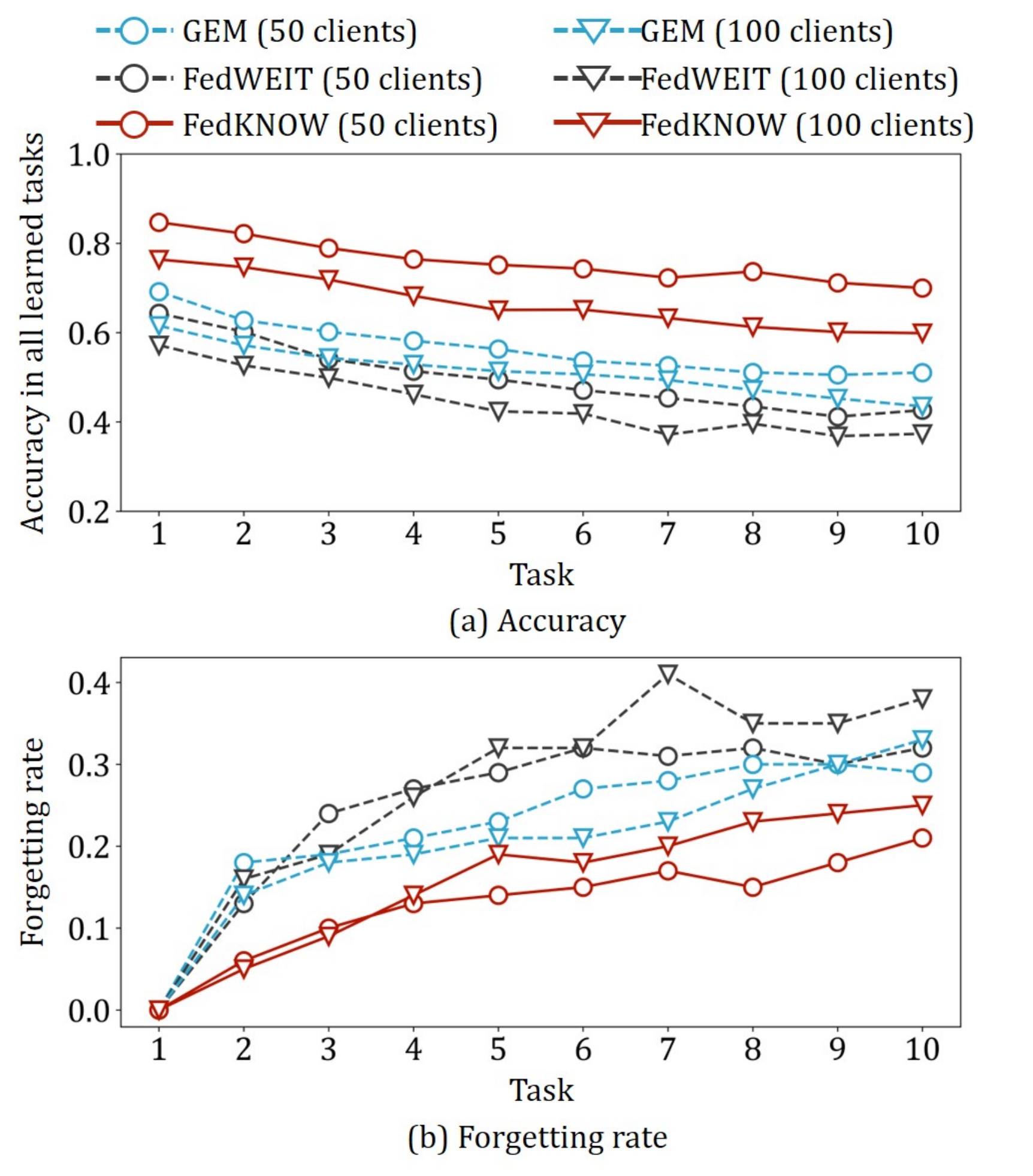}\\
  \caption{\textcolor{black}{Discussion of model accuracies under different numbers of clients}}
  \label{Fig: DiscussClient}
\end{figure}

\textbf{Number of clients}. Two cluster scales are considered in this evaluations: 50 and 100 clients. This is because when distributing the MiniImageNet dataset to these clients. Each client only has a small number of training samples when the client number is 100.
Figure \ref{Fig: DiscussClient} displays the comparison of three methods.
We can see that for both client numbers, FedKNOW can achieve the highest accuracies (Figure \ref{Fig: DiscussClient}(a)) and our approach has the lowest forgetting rates (Figure \ref{Fig: DiscussClient}(b)). This is because when the client number becomes larger and the training samples among different clients become more heterogeneous, the negative knowledge transfer becomes more severe. Our approach optimizes the gradient integration process and tries to minimize the influence of such negative transfer when learning new tasks, thus gaining more accuracy improvement when the degree of data Non-IID increases.
In most of the cases, GEM has higher accuracies than FedWEIT. This is because FedWEIT decomposes the weights of each layer into adaptive and base ones. However, ResNet-18 has several downsample layers with few but important weights, and the decomposition these layers degrades the model accuracy. In contrast, GEM maintains the architecture of the model and uses retained samples that provide higher accuracies than FedWEIT.

\subsection{Applicability in Different DNNs and Settings} \label{Sec: Discussions}


\textbf{Applicability of FedKNOW to DNNs}. FedKNOW represents the first framework that supports knowledge-level federated continual learning of DNNs for edge-based environment. FedKNOW can be generalized to support most of state-of-the-art DNNs~\cite{khan2020survey}.
To support this claim, we implemented and tested 8 DNNs beloning to six typical categories: (1) depth (ResNet-152~\cite{resnet}); (2) multi-path (DenseNet~\cite{densenet});
(3) width (InceptionV3~\cite{inceptionv3}, ResNeXt~\cite{resnext}, and WideResNet~\cite{wideresnet});
(4) feature map exploitation (SENet18~\cite{senet});
(5) attention (ResNet-152~\cite{cbam} and SENet18~\cite{senet});
and (6) lightweight DNN (MobileNetV2 whose width multiplier is 1.0 and width multiplier is 2.0~\cite{mobilenetv2} and ShuffleNetV2~\cite{shufflenetv2}). The evaluation settings of these models follow Section \ref{Sec: Effectiveness}.

\begin{figure*}[htp]
\centering
  \includegraphics[scale=0.52]{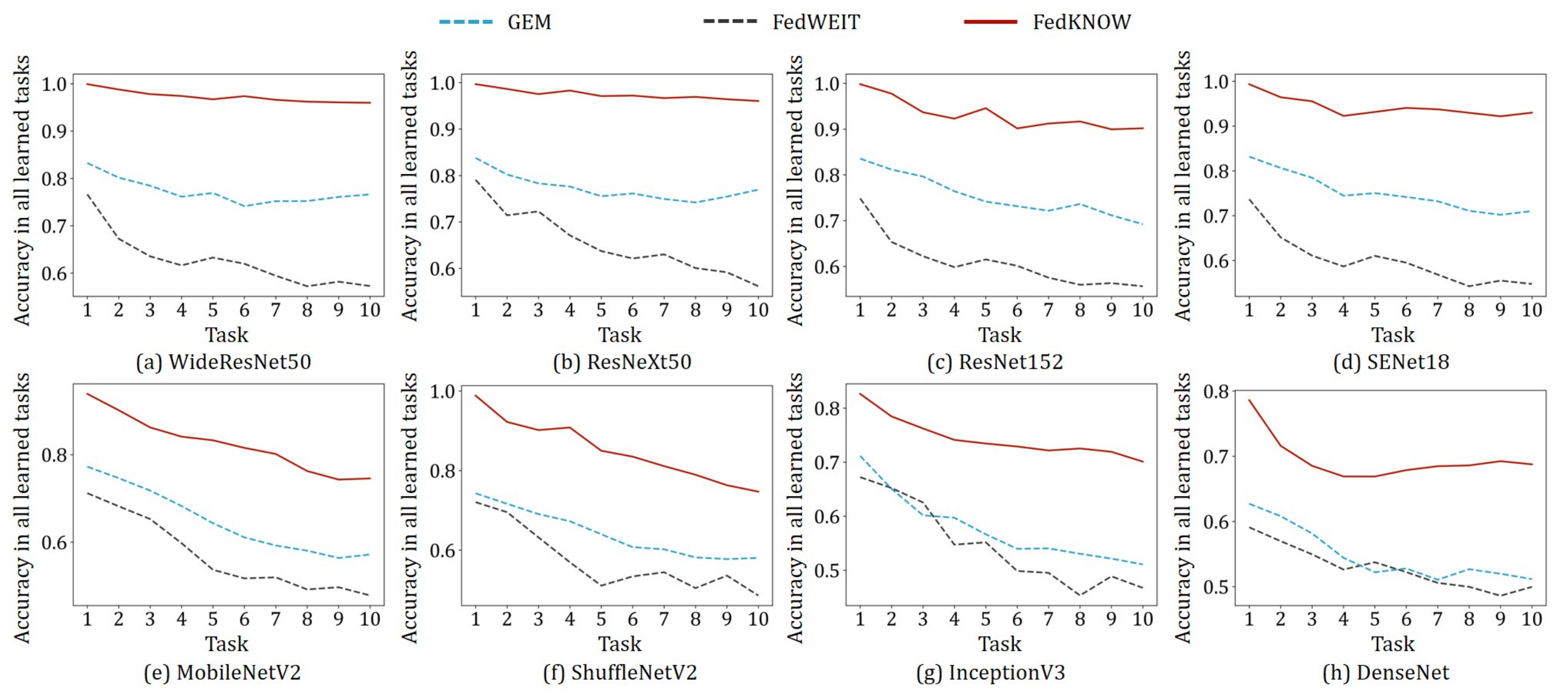}\\
  \caption{Applicability of FedKNOW to six categories of DNNs}
   \label{Fig: 8DNNs}
\end{figure*}

Figure \ref{Fig: 8DNNs} illustrates the comparison results when applying GEM, FedWEIT, and FedKNOW to learn the 10 tasks in MiniImageNet. The results show that when a new task comes, our method can re-train all these DNN models to maintain high accuracies for all learned tasks. Although the accuracies in all three methods decrease when the task number increases, our method provides the highest accuracies in all cases. This is because our knowledge extraction and integration mechanism maintains and restores the most task knowledge in the continual learning process.
In addition, we can see different models have different generalization capabilities and thus have different accuracies when learning the same tasks. For example, when comparing to the ResNet-18 model used in previous evaluations (Figure \ref{Fig: ComparisonBasic}(d)), small models such as MobileNetV2 have lower accuracies (Figure \ref{Fig: 8DNNs}(e)), while the large models such as WideResNet50, ResNeXt50, ResNet-152, and SENet18 ((Figures \ref{Fig: 8DNNs}(a) to (d))) have considerable higher accuracies. The wide applicability of FedKNOW makes it possible to use large models on small edge devices to deliver high accuracies.

In contrast, FedWEIT has the lowest accuracies when training these models. The main reason comes from the design of its loss function, which consists of three components: (1) the task loss; (2) the sparsity regularization penalty term for task-adaptive weights and masks; (3) the difference between the weights of two consecutive time steps. In training, two hyper-parameters are used to balance these components.
However, in this evaluation, different models have considerably different architectures and numbers of weights (for example, the size of ResNet152 is 117 MB while the size of MobileNetV2 is only 3.5 MB). This means the two hyper-parameters have to be carefully tuned for each model in its training.
Moreover, for compact networks such as MobileNetV2, ShuffleNetV2, and ResNeXt, sparse model weights may have no volume to store much knowledge. This further degrades the accuracy of FedWEIT.

\begin{figure}[htp]
\centering
  \includegraphics[scale=0.51]{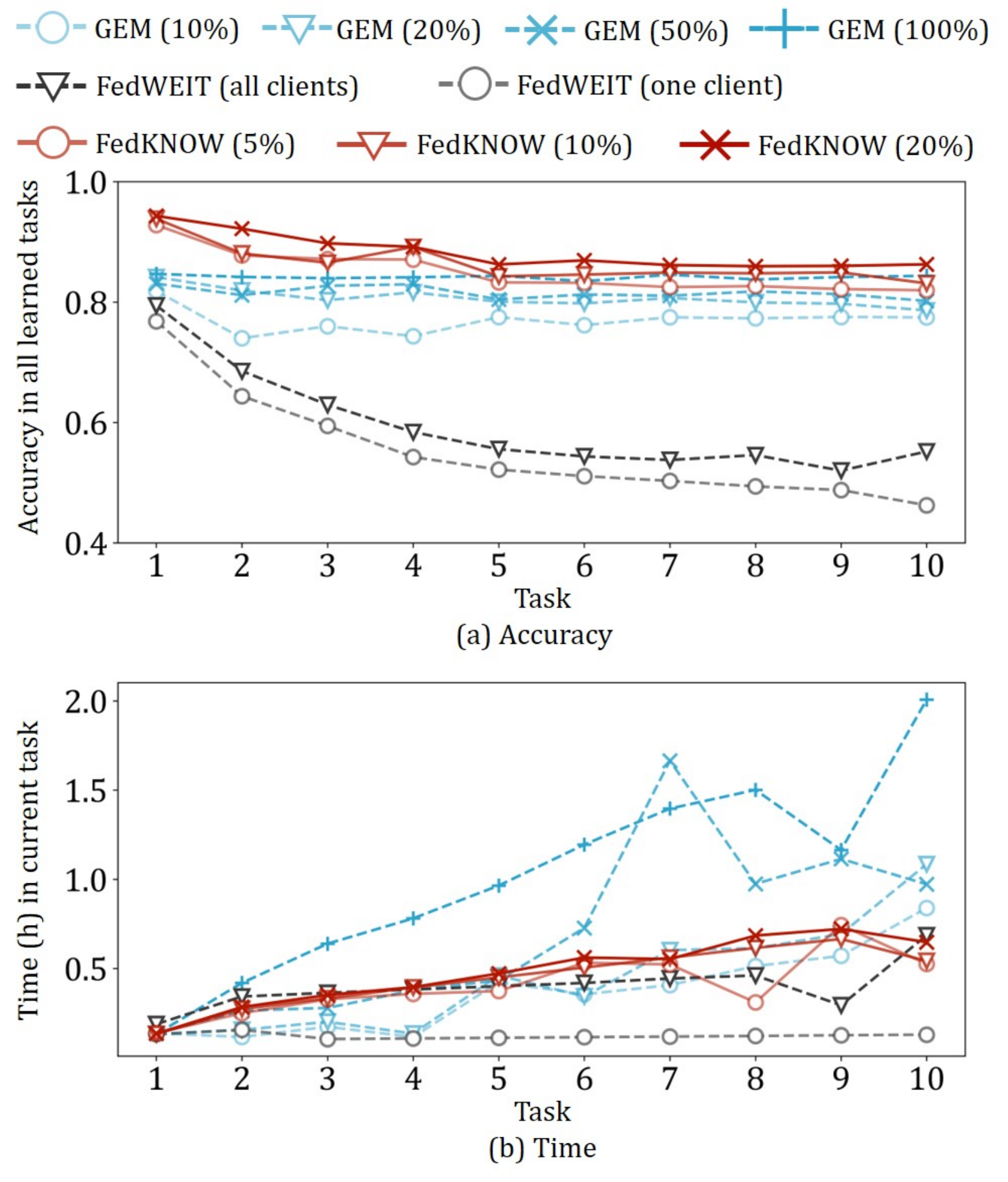}\\
  \caption{Discussion of different parameter settings}
  \label{Fig: ParameterSettings}
\end{figure}

\textbf{Discussion of different parameter settings}. In this evaluation, we discuss three typical methods to retain previous knowledge in continual learning and show the impact of different parameter settings in these methods:
(1) GEM stores 10\%, 20\%, 50\%, 100\% of each task's training samples and uses them when learning new tasks.
(2) FedWEIT decomposes and regularizes adaptive weights from each task's learned model. In evaluation, we consider two settings: FedWEIT uses adaptive weights from all clients (the original setting) and it only uses adaptive weights from its own tasks in one client.
(3) FedKNOW extracts the most important weights of each task as its knowledge. Three percentages of weights with the largest values are tested: 5\%, 10\%, and 20\%. MiniImageNet and ResNet-18 are tested here. Figure \ref{Fig: ParameterSettings} shows the model accuracies and training time of the three methods under different settings and we have two key observations.

\emph{Model accuracy}. The results in Figure \ref{Fig: ParameterSettings}(a) show that retaining less information (samples or model weights) indeed causes lower accuracies in each method.
We can see that even when GEM uses 100\% of previous training samples, it still provides lower accuracies than FedKNOW in most of the cases. This is because GEM does not consider negative knowledge transfer from other clients' Non-IID data.
In contrast, FedKNOW optimizes the gradient integration to minimizes the influence of negative transfer while preventing forgetting previous tasks' knowledge.
This allows our approach to avoid the influence of uncertain knowledge transfer from other clients in real scenarios, in which the clients participating in learning the model may dynamically change over time.

\emph{Model training time}. As expected, Figure \ref{Fig: ParameterSettings}(b) shows retaining and processing more previous information takes longer training time. FedWEIT only using one client's adaptive weights takes the shortest time because it processes the least information, which is insufficiently to maintain previous knowledge and hence this method has the lowest accuracy.
GEM suffers most from processing more training samples. For example, the model training time on 100\% previous samples is 2 to 3 times longer when only 10\$ of previous samples are used.
In contrast, the differences of model training time among three knowledge sizes are much smaller in FedKNOW. This allows our approach to use larger knowledge to increase model accuracy.

\section{Conclusion}


This paper presents the design, implementation and evaluation of FedKNOW, a framework that enables accurate and communication-efficient federated continual learning on distributed edge devices. FedKNOW is based on a novel optimization to integrate signature task knowledge. Our approach extracts and retains task knowledge from all learned tasks, while optimally assembling the most important knowledge to adapt to a new task with high accuracy and low communication overheads. Extensive evaluations in real scenarios against latest federated continual learning techniques strongly prove the efficacy and practicality of FedKNOW, especially for challenging learning scenarios of 80 different tasks and 100 clients.


\bibliographystyle{plain}
\bibliography{references}

\end{document}